\RequirePackage{amsmath}
\documentclass{llncs}

\usepackage{srcltx} 
\usepackage{times,latexsym,array,verbatim,url}
\usepackage{epsfig,color,graphicx,cite,epstopdf}
\usepackage{mdwmath,mdwtab,mathptmx}
\usepackage{wrapfig}

\usepackage{xspace} 

\begin{document}
\frontmatter          

\mainmatter              
\title{Toward the Starting Line:  \\ A Systems Engineering Approach to Strong AI}
\titlerunning{Toward the Starting Line}  
%
\author{Tansu Alpcan \inst{1}
\and Sarah M.~Erfani \inst{2}
\and Christopher Leckie \inst{2}
}
%
%
%
\institute{Dept. of Electrical and Electronic Engineering, The University of Melbourne, VIC, Australia.\\
 \email{tansu.alpcan@unimelb.edu.au}
\and
 School of Computing and Information Systems, The University of Melbourne, VIC, Australia.
}

\maketitle              

\begin{abstract}
Artificial General Intelligence (AGI) or Strong AI aims to create machines with human-like or human-level intelligence, which is still a very ambitious goal when compared to the existing computing and AI systems.
After many hype cycles and lessons from AI history, it is clear that a big conceptual leap is
needed for crossing the starting line to kick-start mainstream AGI research. This position paper aims to 
make a small conceptual contribution toward reaching that starting line.
After a broad analysis of the AGI problem from different perspectives, a system-theoretic and engineering-based research approach is introduced, which builds upon the existing mainstream AI and systems foundations.
Several promising cross-fertilization opportunities between systems disciplines
and AI research are identified. Specific potential research directions are
discussed.

%
%
\end{abstract}

%

\section{Introduction} \label{sec:intro}

The terms \textit{Strong AI} or \textit{Artificial General Intelligence} (AGI) refer to the type of AI that has
certain capabilities similar to those of human intelligence.  The AGI concept
differs significantly from the existing computing systems and mainstream AI algorithms also called as ``weak AI''.
Weak AI systems are very successful in solving specific problems when the problem context is provided directly by a human programmer. In contrast, AGI is much deeper and broader in its ambition as it aims for human-like flexibility, understanding, and creativity \cite{kurzweilbook}. Even though it is not well-defined, AGI is one of those things that ``we will know it when we see it'' \cite{knowitseeit}.

The fact that we do not have machines with human- or animal-level intelligence today is indisputable. Current
computers run manually programmed or trained algorithms tailored to fulfill specific functions, whereas
animals and humans automatically adapt to and manipulate their environments. Moreover, the view that the \textit{Strong (or Hard) AI problem} has not yet been well defined \cite{hutterbook} is shared by many experts. Maybe this should not come as a surprise given that we don't have a solid grasp on what  human intelligence means. The confusion on what AI means and can achieve has existed since the beginning of the field in 1950s \cite{aihistory1,buchanan2005very,Crevierbook}. The history of AI consists of one hype cycle following another, in which the pendulum has been swinging between wildly optimistic predictions \cite{kurzweilbook} and deep pessimism \cite{dreyfus1}.

The active research fields on \textit{weak AI} such as machine and deep learning (DL), pattern recognition, and data sciences are currently very healthy and growing after a long AI winter. In contrast, after disappointments of many hype cycles, the dream of creating AGI is currently sustained only by a small set of idealists \cite{agiconf,kurzweilbook,buchanan2005very}. We are today arguably behind the \textit{starting line}, which needs to be passed for the AGI research to become mainstream.

Systems theory and engineering encompassing control, distributed systems, information, and
communications theories as well as signal processing disciplines can potentially play a significant role
in moving toward the starting line of AGI research. Many AI research fields and systems theory share the same
mathematical underpinnings and aim to achieve similar goals in different ways. Therefore,
a cross-fertilization, which has been hindered mainly by academic inertia, may become
very productive.

The goal of this speculative paper is to make a small conceptual contribution by advocating a
systems approach to AI. The next section analyses the AGI problem from different perspectives and studies
the questions of what do we want from AGI, how successes of mainstream
AI of today lay the foundations for tomorrow, what are the different types of (artificial) intelligences,
and why are we (still) behind the starting line when it comes to AGI.
Section~\ref{sec:sysengai} advocates a systems approach to AI incorporating the biological,
information and data, and distributed systems/networking dimensions. The paper concludes with Section~\ref{sec:vision}, which introduces a specific research vision for AGI building upon existing work and discusses the potential role of systems theory and engineering for AGI research.

%
%
%

\section{AGI Problem Analysis: Behind the Starting Line} \label{sec:analysis}

A proper analysis of the AGI problem would span multiple books due to its complex multifaceted
and multi-disciplined nature. This section aims to discuss only certain aspects of the problem
as a precursor to the systems approach presented in Section~\ref{sec:sysengai}.

\subsection{What do we want from AGI?} \label{sec:strongai}

The answer to the question of what we want from (Strong) AI has always been simple to state:
\textit{machines with human-like intelligence} \cite{kurzweilbook}. It is rare that such a simple statement can be so ill-defined.
Concepts such as creativity, flexibility, autonomy are often used to clarify what is meant by
human-like intelligence and the topic has been the subject of many philosophical discussions.
Two observations are worth making. Firstly, we really don't know much about how human (or animal) brain works, so we have a limited scientific understanding of human-like intelligence \cite{nihbrain}. While an engineered
AI may end up looking very different from biological brain, the underlying principles have to be discovered and digested. The planes and birds fly differently, yet flying is flying when it comes to aerodynamics.
Secondly, even with our limited knowledge, it is easy to argue that the current computing systems  are as far
from human-like intelligence as it gets in terms of these aspects. Therefore, we can argue that we are behind the starting line of building machines with human-like intelligence.

Another, maybe a more practical answer to the question is: \textit{more than what we have now} (in mainstream AI). Despite the recent great successes of mainstream AI research, there are multiple core problems that are considered AI-complete or AI-hard \cite{shapiroai}. We can count fields of computer vision and natural language understanding among these.
When we ask for human-like conceptual understanding, this is beyond the capability of current machines \cite{dreyfus1},
even when their closely human-supervised algorithms perform great in specific tasks \cite{hsudeepblue, pokeralberta1}. Another interesting AI-complete problem is networked system security \cite{alpcanbook}, which will remain an open issue as long as computing systems lack autonomy and certain degree of human-like awareness since a tool can always be misused by malicious parties who can access it.

Beyond what we want, there are two additional questions that have been hotly debated. The first one is \textit{whether we should want AGI}. The AI field has never lacked optimists. Recently, there has been exaggerated
claims (again) on `how thinking machines will take over the world soon' \cite{kurzweilbook}, which turned into a
borderline scare campaign \cite{bostrom2014superintelligence} that influenced many non-experts in the broader society. The arguments in this paper show that there is no need to panic when it comes to AGI progress.

The second question is \textit{whether (human-created) AGI is even feasible}. The field never lacked pessimists either. Maybe the most famous of them has been Hubert Dreyfus, who has contributed to the field of AI with well-thought (but perhaps destructive) criticism \cite{dreyfus1}. At the risk of oversimplification, his main arguments center around the fact that the underpinnings of human intelligence is non-symbolic, which follows the insights of philosophers such as Heiddeger \cite{dreyfus1991being}. Unsurprisingly, his views have met with hostility when symbolists were dominant in AI history, but have been received more positively by connectionists (deep, neural, and statistical learning researchers). On the flip side, the well-known earlier work by Minsky and Papert \cite{minsky1972perceptrons} ``showed'' the perceived limits of connectionism and instead has given the symbolic approach an upper hand in the field for a while. It turned out a decade or so later that their conclusions were quite incomplete and those perceived limits have been addressed through subsequent discoveries in neural networks (NNs) \cite{rumelhart1985learning}. Some say that the work even contributed to the so-called AI winter. Throughout the AI history, the pendulum has been swinging between exaggerated optimism and deep negativity. This can also be interpreted as a sign of the field's immaturity.

In summary, the feasibility question is wide open and will probably remain so until we have a clear analytical or mathematical understanding of what we really want. For example, are AGI goals achievable within the Turing Machine formalism? If not what extensions are needed? Even if it is feasible theoretically, the computational resources needed may be prohibitively high, especially considering that human brain has 100 billion neurons. Some of the recent theoretical work sidestep the latter issue by (perhaps optimistically) assuming future availability of  unlimited computational resources \cite{hutterbook}.

\subsection{Successes of Mainstream AI} \label{sec:weakai}

The current state-of-the-art AI research has been experiencing a boom
within the last decade. The impressive successes of mainstream AI arguably lay the foundations of AGI but
may not be sufficient as is. The following developments are noteworthy: \\
\textbf{(1)} From an algorithmic perspective, most of the early heuristic approaches have now strong mathematical formulations, which provide solid foundations for future improvements and synthesis \cite{horniknn,siegelmann1997computational,Bishopbook,MacKayGP,Schoelkopfbook}. Several classical AI problems, e.g. playing complex games such as chess, go, and poker have been effectively solved. Significant progress has been made in
 various fields, e.g. image/speech recognition. \\
\textbf{(2)} The mainstream AI research also has a strong experimental culture that helps addressing real world problems.
 Moreover, the field has numerous practically-relevant application domains which easily attract popular attention and research funding. \\
\textbf{(3)} There has been enormous advances in the hardware domain. The advances in computing power (central, graphical,
 and application processing units) as well as direct access memory and storage capacities (disk drives) have been exponential. \\
\textbf{(4)} The exponential improvements in networking (in terms of bandwidth, convenience, and costs) and maturing Internet have resulted in a data revolution, which has boosted data-oriented approaches significantly.



\subsection{Different types of (artificial) intelligences} \label{sec:aitypes}

A basic definition of intelligence is ``the ability to acquire and apply knowledge and skills''~\cite{shanelegg1}. Based
on this definition, humans have been building machines with limited intelligence for a long time.
An adaptive controller such as cruise control in cars or autopilot in planes can be considered ``intelligent''
in the simplest sense. Similarly, carefully designed rule-based methods, machine learning algorithms, and
NNs have been used to construct intelligent systems that acquire limited knowledge (e.g. parameters) and achieve well-defined goals using models directly programmed by human.

Purely to further the discussion, let us consider the following arbitrary categories of intelligence types:
(i) controllers in control systems (ii) programmed automata and Turing machines (iii) statistical and DL algorithms (iv) biological intelligence. The first three types fit the general definition in varying degrees whereas natural biological systems provide the definition itself. This broad range somewhat explains the conflicting opinions about AI; it is a bit like the ancient story ``blind men and the elephant.'' It is easy for different people to
have  varying opinions on the topic based on their  focus.

When it comes to AGI, what we want is clearly closer to the biological type of intelligence. However, we
still know very little about biological computing systems \cite{nihbrain}. 
For example, we still don't know the exact mechanisms of how brain stores information (in short and long terms). We have a very limited understanding of brain's plasticity or organization. We don't know how brain develops
from a single cell in an embryo to enormous complexity in a few years.
A solid understanding of the principles behind biological intelligence is needed to build systems with AGI. 


\subsection{Why are we (still) behind the AGI starting line?} \label{sec:belowzero}

The AGI problem is still not well-defined even after decades of investigation \cite{hutterbook}.
The main issue is not the specific definition or lack of it per se, but
it is that we don't even know how to make the transformative leap despite many past attempts \cite{aihistory1,buchanan2005very}.
The emergence and successes of weak AI, which is the mainstream AI for now, can be attributed to realizing this issue and scaling down the ambitions of AGI. The resulting sensible approach has paid significant dividends and great
progress has been made in numerous  fields ranging from image recognition to automated translation.
The undeniable successes of weak AI in areas which were once considered the domain of AGI such as the Turing test~\cite{mauldin1994chatterbots}, playing full information games, e.g. \textit{chess}  \cite{hsudeepblue},  and \textit{go} (DeepMind)  as well as limited information ones, e.g. \textit{poker} \cite{pokeralberta1}, certainly support the
collegial joke that ``when an AI problem is solved it is not (of interest to) AGI anymore.''

Can the incremental mainstream AI research erode the mountain of AGI? The answer is negative as long as
we don't have a good starting point. The need for the paradigm shift is due to the fact that AGI we want is very different from the existing computing systems, including weak AI algorithms. Once we reach the starting line, however, the incremental progress will start.
The argument that we are (well) behind the starting line is supported by the following observations: \\
\textbf{(1)} We don't have a proper scientific understanding of AGI objectives beyond vague allusions. There are no clear analytical or mathematical descriptions of what is meant by AGI properties such as creativity, flexibility, and autonomy. \\
\textbf{(2)} The AGI research, as of first two decades of the 21st century, is not part of mainstream AI research.
 The few conferences and journals on the topic are considered niche and more philosophical than business-as-usual. \\
\textbf{(3)} Most of the mainstream AI researchers are very realistic about their results. It is not claimed that  existing recognition or game playing systems have a human-like understanding or flexibility of applying the know-how gained to other problems. \\
\textbf{(4)} The theoretical and practical knowledge basis we have falls well short for the usual incremental research progress when it comes to AGI goals. The gap is simply too big between data-oriented (statistical or deep) learning algorithms or symbolic formalisms and the AGI goals. Recent great theoretical accomplishments in fields such as algorithmic information theory or decision theory, despite providing
 pieces of the puzzle, also fall short in paving an obvious way toward human-like AI systems.

These observations and arguments become clearer when we further analyze next what we really want from AGI. Without such an analysis, the collegial joke above becomes quite serious!

\section{A Systems Approach to AI} \label{sec:sysengai}


\subsection{Biological Dimension}

Since AGI aims to imitate properties of biological systems, it is appropriate to start
the discussion with the biological dimension. There are several aspects of biological
systems that are arguably directly relevant to AGI system development~\cite{thecell}.
Biological intelligence is always embodied in the physical world. AGI properties
such as adaptability (flexibility) and autonomy can only be defined within the context of
an environment. This important point has been recognized by AI researchers,
see e.g. embodied cognition, Nouvelle AI, etc. \cite{Brooksembodied}. Even the strip of tape in the
Turing Machine abstraction can be interpreted as an ``environment.'' One of the main points in Dreyfus'
arguments against symbolism, which is disembodied, is based on Heidegger's philosophy where
intelligence is naturally integrated into the environment \cite{dreyfus1991being}.

\begin{figure}[tp]
\centering
 \includegraphics[width=0.6\columnwidth]{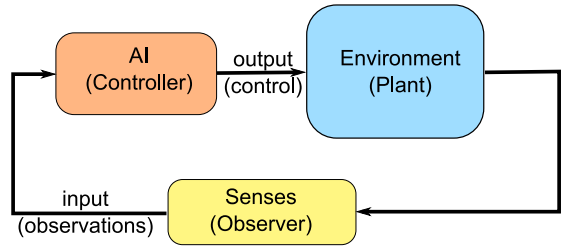}
 \vspace{-2mm}
 \caption{A control-theoretic formalism for AGI systems. \label{fig:controlai}}
\end{figure}
Biology strongly suggest that it is necessary to consider an AGI system
as part  of the environment it inhabits and interacts with. The conceptual
framework used by control system theory~\cite{Anderson1971} to distinguish between a controller (AI), actuator,
dynamical system (environment), and observer, provides a ready-made formalism as shown in Fig.~\ref{fig:controlai}.
The \textit{embodiment of AI} can be but does not have to be physical (e.g. robotics). It is conceivable and sometimes desirable to develop AI systems embedded in virtual environments (e.g. gaming worlds\cite{arcadeaienv}, 3D Internet~\cite{alpcan3dnet}) or other data streams.

Introducing a \textit{Universal Continuity Conjecture (UCC)}, it is possible to consider
inanimate objects, living beings, and intelligence (of all types, including the vague concept of consciousness)
within the same continuous spectrum imposed by the underlying physical laws. In this philosophical view, living organisms as well as different types of intelligence are natural consequences of underlying physical laws.
The fact that we cannot easily draw the boundaries of life and intelligence supports this conjecture. Another strong argument in favor of the UCC is the fact that intelligence builds upon life (cellular organisms) and life builds upon inanimate materials (biochemistry) \cite{thecell} in a seamless way. Despite its purely philosophical nature, UCC can be useful in developing the underlying engineering principles for AGI.

AGI as an embodied system within an environment can be equivalently considered as an organism
with homeostasis (self-regulating) property, consistent with the UCC. Hence, the system approach
goes beyond the computational one and aims to engineer a type of self-regulated embodied organism or complex
system that interacts with the environment. In this sense, the approach builds upon
dual (or adaptive) control~\cite{Wittenmark95adaptivedual} as well as reinforcement learning~\cite{russellnorvigai}.

An illustration of the UCC can be found in the famous Conway's Game of Life~\cite{conway1970game}, which played a significant role in cellular automata research~\cite{Hopcroftautomata}. Conway's Game of Life is a cellular automaton that showcases how abstract objects following simple rules can evolve complex (and stable) structures. While the research on digital organisms and artificial life are  closely related to AGI, the engineering approach adopted here differs from those efforts. The focus is on conscious design and engineering of self-regulating systems instead of waiting for them to emerge through purely evolutionary processes.

Is it possible to achieve the level of flexibility desired in AGI through purely human-driven engineering design?
Is there a middle ground between classical manual design and hands-off emergence of structures through pure evolutionary processes (as in game of life)? The answer to these questions may lie in the under-explored field
of L-systems~\cite{rozenberglsystems}. L-systems are rule-based formal grammars, and hence, are closely related to automata theory, fractals, and iterative and complex systems. They can arguably be used as a starting point to investigate
the concepts discussed.


\subsection{Information and Data Dimension}

\begin{wrapfigure}{l}{0.6\textwidth}
\centering
 \includegraphics[width=0.6\columnwidth]{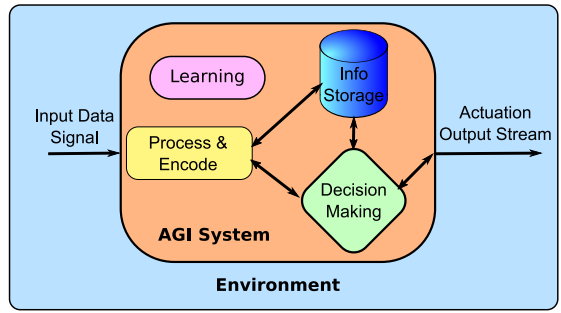}
 \vspace{-2mm}
 \caption{Information processing in an AGI system embodied in an environment. \label{fig:infoai}}
\end{wrapfigure}

An intelligent system embodied in an environment has to process the input data 
from its own sensors, extract the relevant \textit{information} or \textit{knowledge},
and store an encoded version of this knowledge for future usage as shown in Fig.~\ref{fig:infoai}.
An adaptive autonomous system has to incorporate learning into all these functions.
Similar problems have been investigated by reinforcement learning
as well as adaptive and dual control research~\cite{wittenmark1}.

Multiple fundamental system theory fields can potentially play a very useful role in this direction.
Signal processing studies the theory and methodologies of processing the information from a broad range of
signals. Since the AI input data streams (e.g. provided by specific system's sensors which
can be physical or virtual) are signals, the approaches developed in the signal processing
area directly apply. Similarly, data science has recently
emerged and attracted a lot of interest as an interdisciplinary field focusing on
extracting knowledge from data.
Encoding and representation of data has been studied extensively
by coding and information theories~\cite{cover}, which provide an extensive set of mathematical tools.
While these theories have traditionally focused on communication systems, the fundamental
principles can be reused within the model considered.
In conjunction with system-theoretic methods, the database theory provides
a strong foundation for storage of extracted information.

The way information or knowledge is represented and processed has caused a lot of tension
in AI research since the beginning~\cite{papertmanyai}. Connectionist approaches (in the form of NNs and variants) tend to encode information using parameters of NNs, which are trained using
input data, while symbolism adopts a high-level human-readable approach based on
logic (e.g. rule-based or expert systems). Both approaches obviously have useful properties
and provide important insights in their own way.

Design and engineering of AGI systems require building a strong connection between
symbolism and connectionism, while a purely symbolic approach most probably will not provide the autonomy
and flexibility we want from AGI. This is due to the fact that such solutions rely on human-provided
explicit contextual information through modeling and programming.
However, pure connectionism tends to encode knowledge in a form not accessible or interpretable by humans.
The black-box solutions of NNs have been criticized for this since their inception.
Translation of knowledge to human-readable formats is necessary at least to facilitate a human-in-the-loop engineering
design process. Biological systems strongly suggest that such a translation is possible. Human brain is a biological neural network yet our abstract thinking is symbolic.


\subsection{Distributed System and Networking Dimension}

Connectionist approaches to AI are closely related to network science and distributed systems.
NNs and more recent variants under the umbrella of DL are networked systems.
Hence, the distributed system and networking dimension has been a part of AI research since the early days.

The idea that complex solutions emerge from simple interacting components has been fundamental
to artificial life, NNs, and more recently DL. This idea is supported
by observations of biological systems which are almost always cellular. Graph theory provides
strong theoretical foundations to networks. As a related topic, the
emergence of global properties in random graphs have been studied by the percolation theory.

\begin{wrapfigure}{l}{0.5\textwidth}
\centering
 \includegraphics[width=0.5\columnwidth]{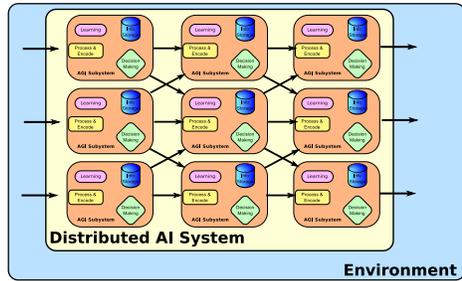}
  \vspace{-2mm}
 \caption{A modular, distributed AGI system architecture inspired by cellular organisms. \label{fig:distrai}}
  \vspace{-3mm}
\end{wrapfigure}

Since the beginning NNs have been inspired by their biological counterparts. It turns out
that the artificial versions we have are overly simplistic. Biological
variants are multi-cellular, where each cell is a (complex) subsystem itself \cite{thecell}. Hence, the natural next
step in connectionism is toward distributed systems consisting of complex subsystems, similar
to multi-core processors, rather than simple functional components (Fig.~\ref{fig:distrai}). 
Recent advances such as Long Short Term Memory (LSTM)~\cite{hochreiterlstm} can be interpreted as developments in this direction. System theory provides a strong foundation for analysis and design of distributed systems building upon know-how from communication networks, distributed optimization, game theory,
and control theory.


\section{Discussion and Future Research Directions} \label{sec:vision}


At the risk of oversimplification, the current mainstream AI research can be crudely clustered into
three main threads that naturally overlap and interleave: (1) symbolic approaches (good old AI), which rely on formal logic and automata theories~\cite{russellnorvigai}, (2) connectionist approaches such as NNs and DL~\cite{schmidhuber2015deep}, (3) statistical pattern recognition and machine learning~\cite{Bishopbook}.
It is natural to expect that these threads will provide some of the building blocks of future AGI systems.

The AI field is again climbing a new hype cycle after a long (in Internet time) AI winter. This cycle
is commercially supported by the Internet-age giants (companies) such as
Google, Facebook, Amazon, and Huawei.
The successes of pattern recognition and machine learning in solving specific problems
as well as exciting recent developments in the DL field exploring new memory and computational architectures~\cite{NTMgraves,Neuralprog,humanleveldeep,memorynets} 
provide a fertile ground for making progress in AGI toward the starting line.
In addition to these data-oriented learning approaches, there has been significant progress in symbolic AI, especially in formal verification methods. The prevalence of standardized data sets for specific problems and the emergence of frameworks for embodiment of agents in physical (robotics) and virtual environments (gaming) provide the indispensable experimental setups and verification.

The mathematical understanding supporting mainstream AI has been maturing. Automata theory, formal languages
and formal verification methods are directly connected to well established discrete mathematics and formal logic
foundations. Most of the existing pattern recognition methods including kernel-based approaches~\cite{Schoelkopfbook} have been formalized using optimization theory. Likewise, NNs and DL~\cite{schmidhuber2015deep} approaches rely on variants of well-known (global and stochastic) optimization~\cite{boydbook,globaloptsurvey} algorithms.

Certain system-theory methods have also been used in mainstream AI research, albeit in a more limited sense.
In DL, Recurrent NNs~\cite{sutskeverrnn} and LSTM~\cite{hochreiterlstm} have been formalized using nonlinear dynamical system models. Entropy-based learning methods have been investigated in the learning area~\cite{schraudolphinfornn}. Game theory has been used to model attacker vs defender interaction in adversarial learning setups~\cite{alpcanbook}.
The link between adaptive control and reinforcement learning has been recently mentioned \cite{hutterbook}.
Morover, NNs can be used for identification and control
of dynamical systems~\cite{Narendranndynamic}.


System theory encompassing signal processing, communications (information theory)
and (distributed) control theories can potentially play a significant role in AGI research.
Most of the mathematical foundations are shared by the AI and system research fields, such as
optimization, graph, game, dynamical system theories as well as discrete event and hybrid systems, and
automata theory. The model-based emphasis of system theory research can sometimes be a restrictive factor yet it can bring the much-needed contextual leap to purely data-oriented (black-box) approaches of machine and DL.
Furthermore, distributed system and game theory can provide a mathematical framework for analysis and development of novel AI systems by rigorously studying individual system components, their environment, and the interactions.

Specific open research questions and directions (in no particular order) include: \\
\textbf{(1)} Combining connectionism and symbolism is necessary to achieve practically useful hybrid and multi-layer system architectures for AGI. This is closely related to how we organize (encode and interpret) the data and learned functions inside the system. \\
\textbf{(2)} Turing machines that can do dual control or reinforcement learning (in an environment)
 can be developed by combining the lessons from control/system theory, building upon recent exciting results in DL such as Neural Turing Machines~\cite{NTMgraves} and Neural Programmer~\cite{Neuralprog}. Better use of fields such as optimization, information, control, and game theories within the context of AI research may result in
 faster progress overcoming the historical unnecessary disconnect between these research communities. \\
\textbf{(3)} Better understanding of biological systems and intelligence is potentially the key to make the leap
 towards AGI. Cellular architectures inherent to living intelligent organisms can be modeled using
 distributed system theory. Specific research subtopics include:
 (a) Analysis and design of ``developing systems'', 
 which evolve from scratch only based on the inherent rules and the environment they are embodied, may give
 clues for the desired adaptability and flexibility. (b) Creation of engineering architectures based on biological principles of homeostasis and cellular organisms would be useful for design of distributed intelligent systems. \\
\textbf{(4)} Addressing the issue of combining explicit human intervention (classical programming) with learning systems
 would help shorten the evolutionary  learning time-scales and achieve backwards compatibility with existing computers.

\section*{Acknowledgements}
The authors thank anonymous reviewers and TPC chairs of the AGI 2017 conference for their valuable
feedback and suggestions.



\end{document}